\documentclass[conference]{IEEEtran}
\usepackage{cite}
\usepackage{graphicx}
\usepackage{float}
\usepackage{amsmath}
\usepackage{listings}
\usepackage{xcolor}
\usepackage{subfig}
\usepackage{url}

\usepackage{tikz}
\usepackage{textcomp}
\usepackage{hyperref}
\usepackage{lipsum}

\newcommand\copyrighttext{%
  \footnotesize \textcopyright © 2020 IEEE. Personal use of this material is permitted. 
    Permission from IEEE must be obtained for all other uses, in any current or future media, 
    including reprinting/republishing this material for advertising or promotional purposes, 
    creating new collective works, for resale or redistribution to servers or lists, 
    or reuse of any copyrighted component of this work in other works.
  DOI: \href{https://doi.org/10.1109/ICCCNT49239.2020.9225571}{10.1109/ICCCNT49239.2020.9225571}}
\newcommand\copyrightnotice{%
\begin{tikzpicture}[remember picture,overlay]
\node[anchor=south,yshift=10pt] at (current page.south) {\fbox{\parbox{\dimexpr\textwidth-\fboxsep-\fboxrule\relax}{\copyrighttext}}};
\end{tikzpicture}%
}

\DeclareRobustCommand{\rchi}{{\mathpalette\irchi\relax}}
\newcommand{\irchi}[2]{\raisebox{\depth}{$#1\chi$}} 

\title{Detection of Customer Interested Garments\\
in Surveillance Video using Computer Vision}
\author{
	
\IEEEauthorblockN{Dr. Earnest Paul Ijjina\IEEEauthorrefmark{1}}
\IEEEauthorblockA{Assistant Professor, Department of Computer Science and Engineering,\\
	National Institute of Technology Warangal, Telangana, India-506004\\
	Email : \IEEEauthorrefmark{1}iep@nitw.ac.in}

\IEEEauthorblockN{Aniruddha Srinivas Joshi\IEEEauthorrefmark{2}, Goutham Kanahasabai\IEEEauthorrefmark{3}}
\IEEEauthorblockA{B.Tech., Department of Computer Science and Engineering,\\
	National Institute of Technology Warangal, Telangana, India-506004\\
	Email : \IEEEauthorrefmark{2}aniruddha980@gmail.com, 
\IEEEauthorrefmark{3}gauthamkanags@gmail.com}
}

\begin{document}
\maketitle
\copyrightnotice

\begin{abstract}
One of the basic requirements of humans is clothing and this approach aims to identify the garments selected by customer during shopping, from surveillance video. The existing approaches to detect garments were developed on western wear using datasets of western clothing. They do not address Indian garments due to the increased complexity. In this work, we propose a computer vision based framework to address this problem through video surveillance. The proposed framework uses the Mixture of Gaussians background subtraction algorithm to identify the foreground present in a video frame. The visual information present in this foreground is analysed using computer vision techniques such as image segmentation to detect the various garments, the customer is interested in. The framework was tested on a dataset, that comprises of CCTV videos from a garments store. When presented with raw surveillance footage, the proposed framework demonstrated its effectiveness in detecting the interest of customer in choosing their garments by achieving a high precision and recall.
\end{abstract}
\begin{IEEEkeywords}
Surveillance Video, Garments detection, Computer Vision
\vspace{-1mm}
\end{IEEEkeywords}
\section{Introduction}
The task of recognition of garments the customers are interested in, is a prerequisite to computer vision applications like customer behaviour analysis and sales forecasting. This domain has attracted a lot of attention due to its application in attracting customers and for retail sales analytics. To develop a garments recognition system for real-world surveillance videos is a challenging task, due to the myriad of sub-tasks involved in analysing the video, which includes the segmentation of garments, face detection and tracking, etc. The automated identification of the garments and their categorization is a challenging task due to the wide range of garments material, colours and models; which is a difficult task even for human, when using a typical indoor single-view surveillance video.\par

The major factors impacting the effectiveness of garments recognition system under real-world condition are a) the deformation and occlusion of clothes being recognized and b) the variation in 
fabric, colour, texture, shape, style, etc., of the garments, which tend to confuse existing models.\par

\section{Related Works and Literature}

One of the early attempts to fashion identification utilizing image segmentation was proposed by Yamaguchi \textit{et al.} in garments \cite{ref1}, where Fashionista dataset was introduced, comprising of approximately 158,000 images of fashion apparel. This approach utilised a fusion of pose estimation and local features to perform semantic segmentation of the said fashion image. In Yamaguchi \textit{et al.} \cite{ref2}, the earlier approach is extended to utilize a data drive learning model based on semantic segmentation. Along similar lines, Hasan \textit{et al.} \cite{ref3}, proposed a segmentation approach to assign a unique label from classes \{Tie, Shirt, Pants\} to every individual pixel in the image. This method primarily focused on datasets that involved people wearing the garment.\par

A deep learning based approach was proposed by Brian Lao \textit{et al.} that utilised a R-CNN caffe model to identify clothing \cite{ref4}, which was evaluated for identifying shirts. Due to the deformable nature of clothes, recognition algorithms often find it difficult to identify garments and clothes in their numerous deformities. Kota Hara \textit{et al.} proposed a deep CNN algorithm that amalgamated background information of human postures to address this task \cite{ref5}. In this approach, the variability of the garments in terms of rigidity is considered, however the coordinates of the object to be detected is relatively close to the detected human posture. This approach relies on human posture detection algorithms for its accuracy, thereby increasing the average detection time, and in turn making it unsuitable for real-time detection systems.\par

Cheng \textit{et al.} proposed a method which utilised a clothing search approach that requires the details of the region of the clothes, thereby generating a virtual assistant to navigate clothes in line with the end user's input \cite{ref10}. The user provides keywords such as formal, casual, etc., and the type of occasion such as date etc., to the model. The recognition module utilises a neural network to identifies appropriate clothes for a date, thereby acting as a personal fashion advisor.\par

Yasuyo Kıta \textit{et al.} proposed a deformable model-driven approach \cite{ref11} to identify fashion garments that are on clothes hangers. The 3D location and orientation of the clothes is considered in the system to identify the discriminative regions of the clothes, to obtain the 3D details of the clothes using the deformable model. In \cite{ref12}, Yasuyo Kita \textit{et al.} proposed a methodology for automatic handling of clothing using visual recognition. Yao \textit{et al.} proposed a method that automatically segments and re-colour garments in an image sequence \cite{ref13}. A Gaussian mixture model is used to identify the background and foreground, followed by a face detection approach to remove those regions in the segmented foreground. A histogram of the remaining pixels is used to model a probability distribution of the clothing chromaticity, which is then repeated for each frame, to obtain the precise region of the clothes using back projection.\par

The approaches discussed so far require high resolution images of garments for effective detection and is therefore not effective for the majority of real-world clothes (or) garments store surveillance video, captured by low resolution cameras at different angles of view, numerous occlusions, etc., which affects the overall performance.

In this work, we prioritise on a high-level classification of garments into three categories. First are the simple clothing items that are monochrome and are more or less deformable. These are easier to be picked up by existing models and hence we focus more on the succeeding two categories. Second is clothing goods that are composed of multiple colours, conventionally Indian apparel that does not have a large dataset with annotated data to train on. This poses challenges to current models as they are harder to be identified due to the seamless transition between multiple colours. Third is the garments that tend to have numerous colours. They are more complex in the sense that they possess numerous deformations that are not feasible to be displayed in its entirety. We aim to address such complex garments in this study.\par

In our work, we have developed an approach to identify the garments the customers are interested in, from the surveillance footage of a small Indian garments store. This footage contains videos having a resolution of 944$\times$576 pixels and have issues such as noise, FPS drops and inconsistent lighting. This dataset makes our work more challenging, but also has an added merit that this approach which performs well under these conditions can be applied to retail garments stores which are present in large numbers in India. The reminder of the paper is organized to present the proposed approach in Section \ref{prop}, the experimental study with results \& analysis in Section \ref{study} followed by conclusions \& future work in Section \ref{conc}.

\section{Proposed Approach}\label{prop}
\begin{figure} 
	\centering
	\includegraphics[width=90mm]{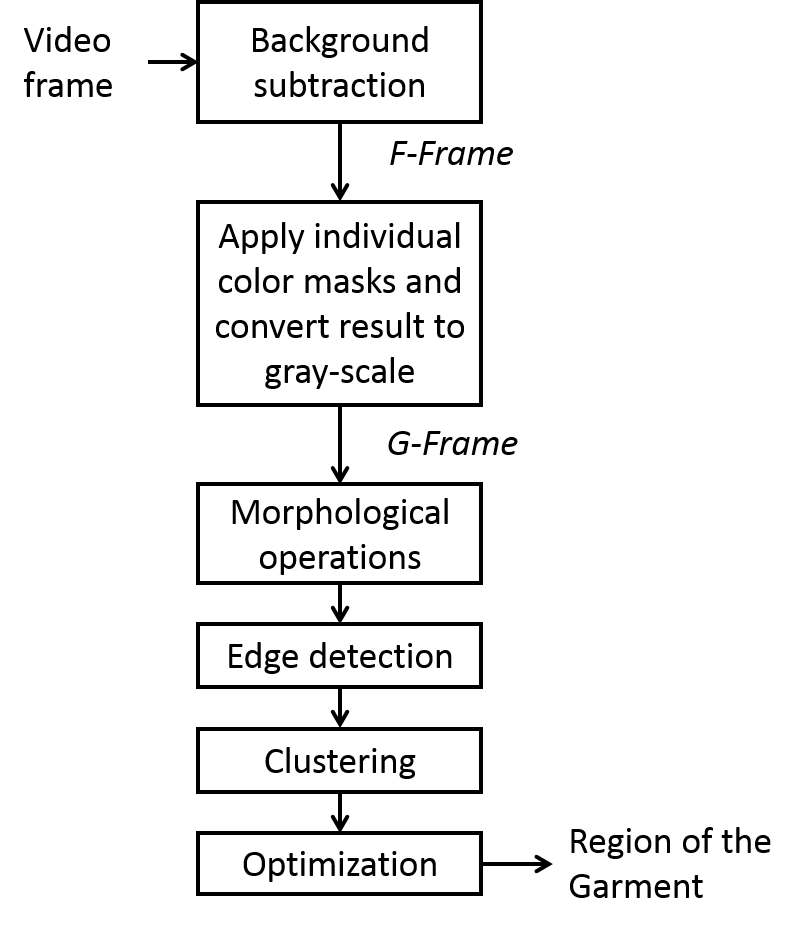}
	\caption{Block diagram of the proposed garments detection approach}
	\label{fig:workflow}
\end{figure}

In this section, we elucidate our proposed framework, which is illustrated in Figure \ref{fig:workflow}. The proposed approach aims to detect garments of interest to a customer from the video surveillance footage of a clothing store.\par
In order to identify the garments of interest, a background subtraction algorithm can be used. The Mixture-of-Gaussian ($MoG$) approach proposed by Zivkovic \cite{ref6} is considered in this work due to its ability to cope with background noise.\par

We now describe the background subtraction process mathematically.\newline
Let $\vec{x}^{(t)}$ denote the value of a pixel in a colour space. Let $BG$ denote the background and $FG$ denote a foreground object.\newline
Let $p(\vec{x}|BG)$ denote the background model. The background model is obtained as a result of estimation from a training set, say $\rchi$. 

Let $\hat{p}(\vec{x}|\rchi, BG)$ denote the estimated model which depends on $\chi$.\newline
The samples in the training set are updated periodically over a chosen time period $T$ to accommodate for the dynamical nature of the scene, such as a new object entering the scene or the exit of an existing object.

At a given time $t$, the training set ${\rchi}_{T}=$  $\{ x^{(t)}, ..., x^{(t-T)} \}$. For every new sample, ${\rchi}_{T}$ is updated and $\hat{p}(\vec{x}|{\rchi}_{T},BG)$ is estimated.
The estimate of foreground objects in the samples of recent history is denoted by
$p(\vec{x}^{(t)}|\rchi\textsubscript{T}, BG + FG)$.

We use a Mixture of Gaussian with $M$ components:
\begin{equation} \label{eq:eq1}
p(\vec{x}^{(t)}|\rchi_{T}, BG + FG) = \sum_{m=1}^{M} \hat{\pi}_{m}N(\hat{x};\hat{\vec{\mu}}_{m}, \sigma^{2}_{m}\text{I}) \end{equation}

where $\hat{\vec{\mu}}_{1}$,..., $\hat{\vec{\mu}}_{M}$ are the mean estimations and $\hat{\sigma}_{1}$,..., $\hat{\sigma}_{M}$ are the variance estimates of the Gaussian components in Equation \ref{eq:eq1}.

The foreground mask obtained from the aforementioned background subtraction algorithm is applied on the given input frame to compute the foreground image (\textit{F-frame}). This \textit{F-frame} is used to compute colour masks corresponding to the major colours i.e., Red, Yellow, Blue, Green, etc. These masks are applied on the given \textit{F-frame} to compute the corresponding images for each colour. Each of these resulting images is converted to a gray-scale image (\textit{G-Frame}). 

\begin{figure}[h]
	\begin{tabular}{c}
	\includegraphics[width=87mm, height=45mm]{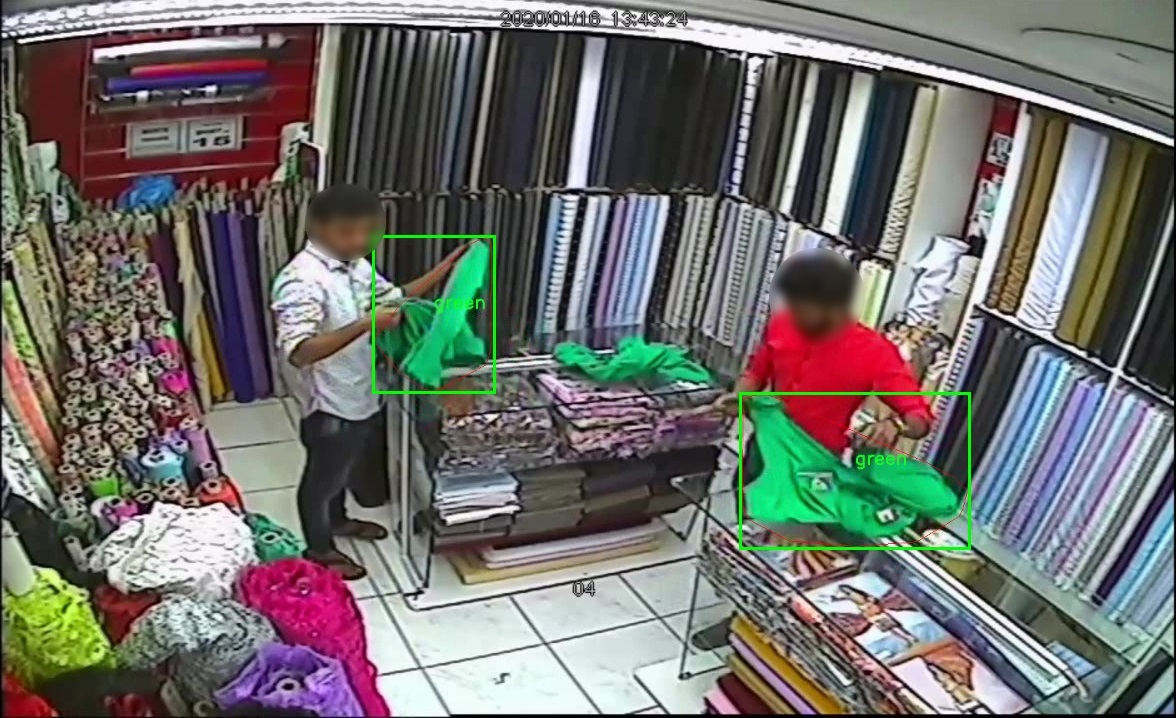}\\
	\includegraphics[width=87mm, height=45mm]{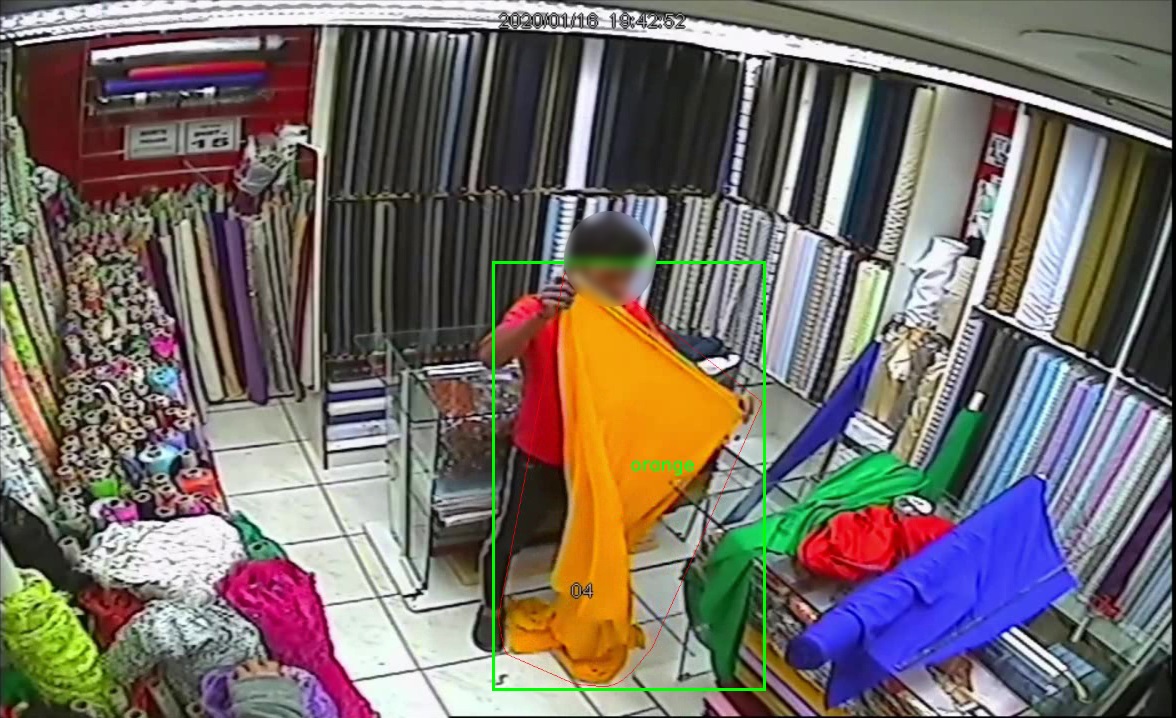}
	\end{tabular}
	\caption{Garments of interest as detected by the proposed approach. (Best viewed in colour)}
	\label{fig:results}
\end{figure}

The morphological \textit{closing} operation is performed on \textit{G-Frame} to remove small holes in the image (due to noise in the video), thereby optimizing the results for this video data. Now, edge detection is used to identify the edges and contours present in the \textit{G-Frame} of each colour. The contours corresponding to small regions are ignored as they might represent patterns on garments with complex embroidery. Finally, clustering is applied to group nearby contours representing the same garments or region. As the customers and sales-person are also captured in the foreground image, further optimization of these clusters is needed. A person detection model such as YOLOv3 \cite{ref7} can be used to identify customers in the frame of interest. Any region present within the bounding box of the detected persons is discarded as these are most likely to be the clothes worn by the customer. The resulting group or cluster of contours are treated as individual garments of interest. The visualization of the garments detected by the proposed approach is shown in Figure \ref{fig:results}.

From the figure, it can be observed that the proposed approach can detect multiple garments the customer is interested in. The object detection model can be further trained to detect only customers. The detected customers can then be mapped to the garments they are interested in using a suitable approach. The garments of interest and customer information can be further utilized to identify the type of garments preferred in a season or by an age-group, and to perform customer analytics for optimizing sales.

\section{Experimental Evaluation}\label{study}
In this section, we describe the dataset created for conducting this study and the results that are obtained by the proposed approach. The experiments were conducted on Google Colab \cite{refx3} with Intel(R) Xeon(R) 2.00 GHz CPU, NVIDIA Tesla T4 GPU, 16 GB GDDR6 VRAM and 13 GB RAM. The program programming environment is Python 3.6 and OpenCV 4.2. The video data used in this study is obtained from a retail sales shop with two surveillance cameras, Stat Vision KC-ID2040D \cite{refx2} capturing the front-view and Keeper CCD Video Camera KC-D3142 \cite{refx1} capturing the side view.

\subsection{Dataset}
\begin{figure*}
	\centering
	\includegraphics[width=180mm, height=175mm]{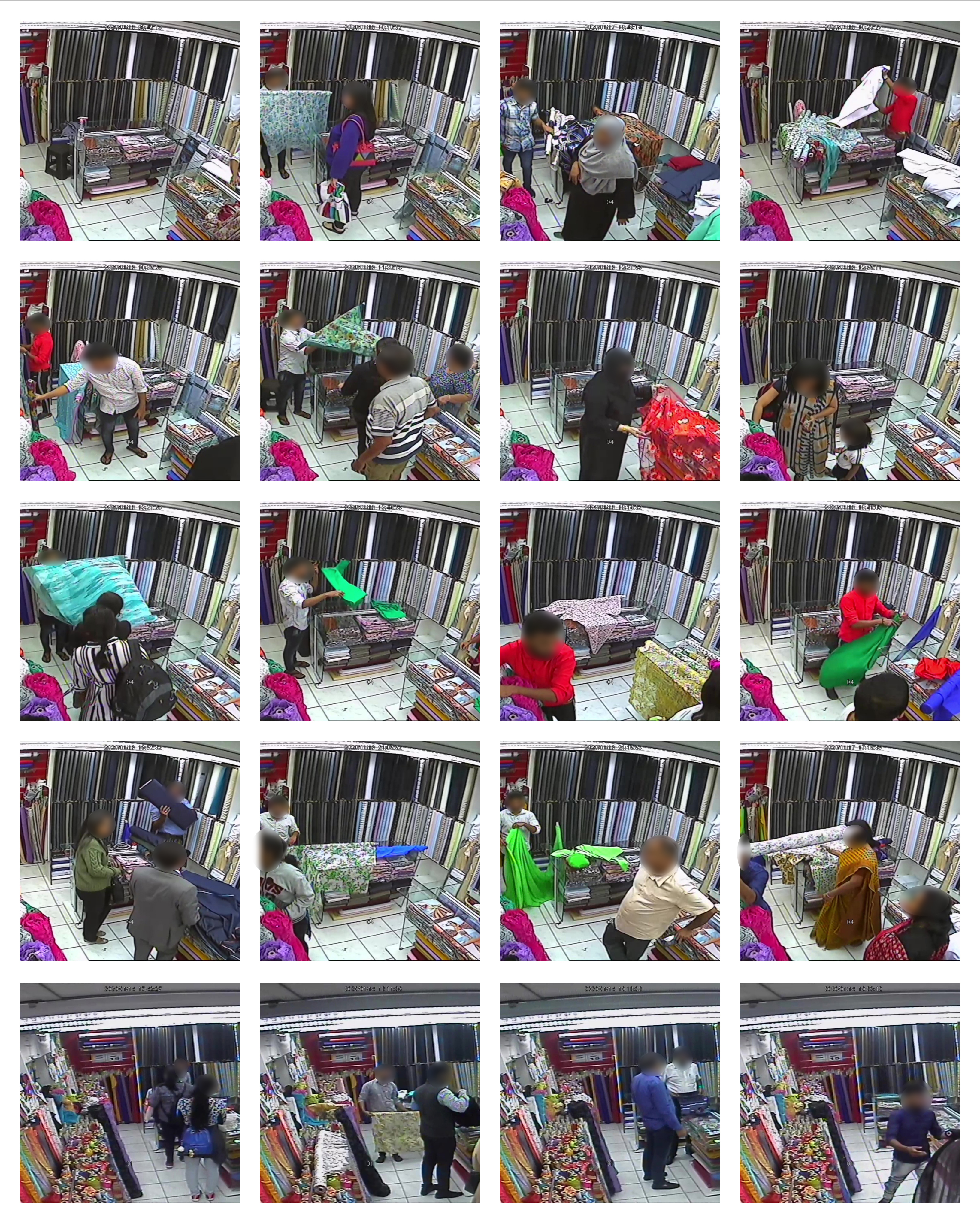}
	\caption{Some of the sales videos in the dataset.}
	\label{fig:dataset}
\end{figure*}

The dataset used in this work is a collection of surveillance videos of an Indian garments store, which mainly comprises of sales information. The videos capture the interactions between the salesmen and their numerous customers throughout the day. The videos are obtained from a total of 2 CCTV cameras (KC-ID2040D and KC-D3142) that are a part of the store's infrastructure, thus enabling us to conduct this study on video captured at different angles of view and illumination conditions. The average resolution of the videos is 944$\times$576 pixels. Raw videos were used without any image enhancing techniques on a per frame basis to stress upon the challenges undertaken in dealing with footage with a considerable amount of background noise. Figure \ref{fig:dataset} illustrates a few sample images from the dataset used for this research.

\subsection{Experimental results and statistics}
The proposed approach has been evaluated using Intersection over Union (\textit{IoU}), a standard evaluation metric used in object detection benchmarks \cite{ref8}. The \textit{IoU} has been used to describe the ground truth of the proposed approach.

Consider $\alpha$ to be the bounding box of the object in interest as obtained from our detection algorithm and let $\beta$ be the bounding box of the same object as annotated in the dataset. Then the \textit{IoU} is measured as:
\begin{equation}
    IoU = \frac{\alpha \cap \beta}{\alpha \cup \beta}
\end{equation}
Consider \textit{TP} to denote the number of true-positive cases, \textit{FP} to denote the number of false-positive cases and \textit{FN} to denote the number of false-negative cases. Then, the \textit{precision} \cite{ref9} which effectively describes the purity of the positive detection relative to the ground truth of the object detection algorithm can be measured as:
\begin{equation}
    Precision = \frac{TP}{TP + FP}
\end{equation}

The \textit{recall} \cite{ref9} which effectively describes the completeness of the positive predictions relative to the ground truth of the object detection algorithm can be measured as:
\begin{equation}
    Recall = \frac{TP}{TP + FN}
\end{equation}

\begin{figure}[h] 
	\centering
	\includegraphics[width=80mm, height=65mm]{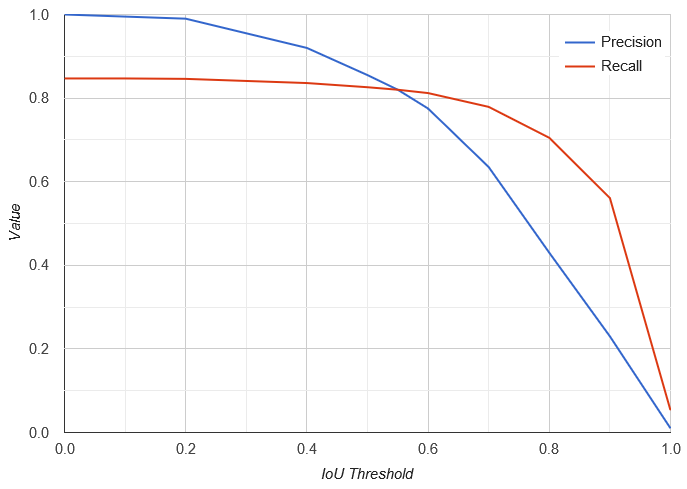}
	\caption{Variation of precision and recall with \textit{IoU} threshold.}
	\label{fig:graph}
\end{figure}

An average \textit{IoU} of 70.81\% was achieved by the proposed approach on the chosen dataset. The variation of \textit{precision} and \textit{recall} with the \textit{IoU} threshold is depicted in the graph shown in Figure \ref{fig:graph}.

\begin{figure}[h] 
	\centering
	\includegraphics[width=87mm, height=55mm]{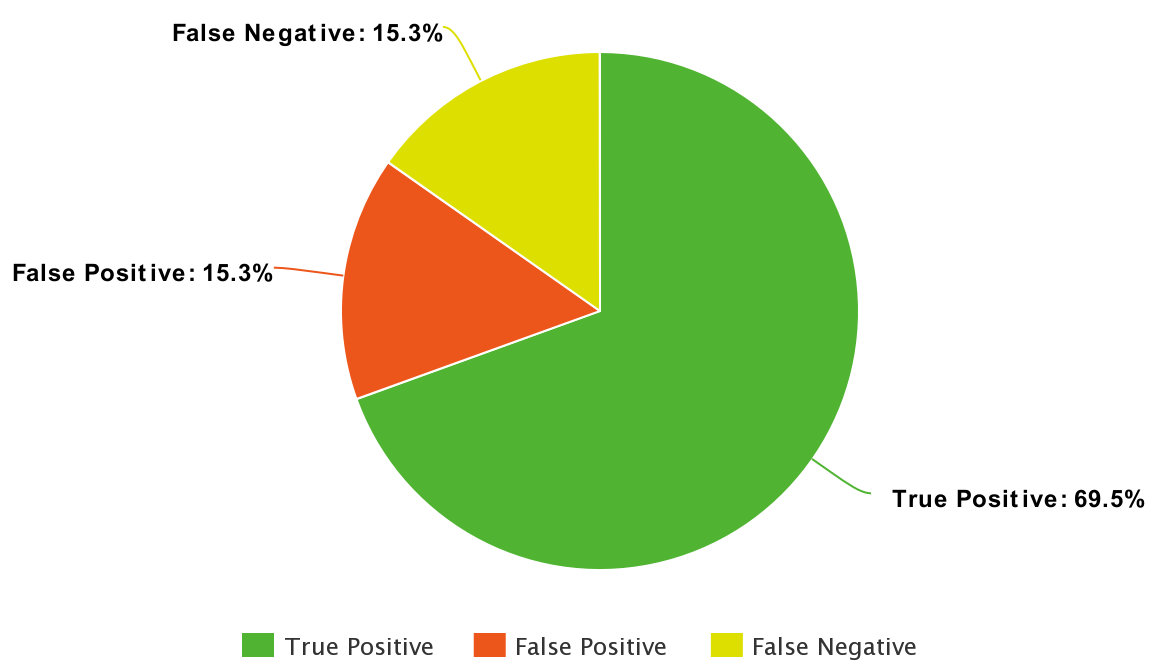}
	\caption{The number of true positives, false positives and false negatives at \textit{IoU} threshold of 0.55.}
	\label{fig:piechart1}
\end{figure}

From Figure \ref{fig:graph}, it can be observed that an \textit{IoU} threshold of 0.55 i.e. 55\% was chosen to distinguish true positive (\textit{TP}) cases from false positive (\textit{FP}) cases and the ratio is depicted in Figure \ref{fig:piechart1}. From this analysis, it can be concluded that the proposed approach has a precision of 0.82 and a recall of 0.82.

\subsection{Analysis of the proposed approach}
From the earlier discussion, it can be observed that the effectiveness of the proposed approach depends on the effectiveness of the background subtraction algorithm and the colour mask. These factors, in turn, affect the \textit{precision} and \textit{recall} of the proposed approach. The \textit{IoU} dropped in instances when only a portion of the garment of interest was detected due to obstruction of the remaining portion by customers during crowded scenarios. The \textit{IoU} was high for garments with consistent colour and which were present in the clear view of the surveillance camera without any obstructions. It is observed that the garments with small strips and chunks of various colours were partially detected and in some cases were not detected by the proposed approach. In some instances, garments were identified as background by the background subtraction algorithm, due to considerable noise in the video frame. These factors influenced the \textit{precision} and \textit{recall} of the proposed approach.\par

\section{Conclusions and Future work}\label{conc}
In this work, a new approach for detecting the garments of interest from surveillance camera video is proposed. The change in visual information during a customer's interaction with the salesmen and the garments is analysed to design an approach for detecting the garments a customer is interested in, using computer vision techniques. The resulting approach is robust and is evaluated on a new dataset proposed in this work. The proposed approach can detect garments of interest with high \textit{precision} and \textit{recall} even in the presence of a considerable amount of background noise and is suitable for cameras having a resolution as low as 944$\times$576 pixels. The \textit{precision} and \textit{recall} are affected by the complexity of the garments being recognized and the density of customers in the shop, which could be the subject of future research in this direction. This work can also be extended to include the analysis of customer interest and to predict the demand for different garments.

\section{Acknowledgement}
This work was done by Mr. Aniruddha Srinivas Joshi and Mr. Goutham K. (B.Tech students) under the guidance of Dr. Earnest Paul Ijjina (Assistant Professor), in Department of Computer Science and Engineering, National Institute of Technology Warangal, as a part of their final year dissertation. The authors express their gratitude to the CSE department and the Institute, NIT Warangal for their efforts in developing the research environment, where this study was conducted. We also thank Google Colab for providing access to the computational resources used for this study.
\bibliographystyle{IEEEtran}
\bibliography{main}
\end{document}